# High-Contrast Color-Stripe Pattern for Rapid Structured-Light Range Imaging

Changsoo Je[1], Sang Wook Lee[1], and Rae-Hong Park[2]

[1] Dept. of Media Technology    [2] Dept. of Electronic Engineering
Sogang University
Shinsu-dong 1, Mapo-gu, Seoul 121-742, Korea
{Vision, Slee, Rhpark}@sogang.ac.kr
http://vglab.sogang.ac.kr

**Abstract.** For structured-light range imaging, color stripes can be used for increasing the number of distinguishable light patterns compared to binary BW stripes. Therefore, an appropriate use of color patterns can reduce the number of light projections and range imaging is achievable in single video frame or in "one shot". On the other hand, the reliability and range resolution attainable from color stripes is generally lower than those from multiply projected binary BW patterns since color contrast is affected by object color reflectance and ambient light. This paper presents new methods for selecting stripe colors and designing multiple-stripe patterns for "one-shot" and "two-shot" imaging. We show that maximizing color contrast between the stripes in one-shot imaging reduces the ambiguities resulting from colored object surfaces and limitations in sensor/projector resolution. Two-shot imaging adds an extra video frame and maximizes the color contrast between the first and second video frames to diminish the ambiguities even further. Experimental results demonstrate the effectiveness of the presented one-shot and two-shot color-stripe imaging schemes.

## 1 Introduction

Triangulation-based structured lighting is one of the most popular ways of active range sensing and various approaches have been suggested and tested. Recently, interests have been developed in rapid range sensing of moving objects such as cloth, human face and body in one or slightly more video frames, and much attention has been paid to the use of color to increase the number of distinguishable patterns in an effort to decrease the number of structured-light projections required for ranging a scene. This paper discusses the design of stripe patterns and the selection of colors to assign to stripe illumination patterns that minimizes the effects of object surface colors, system noise, nonlinearity and limitations in camera/projector resolution for real-time range imaging in a single video frame ("one-shot") and for near real-time imaging in double video frames ("two-shot").

Among the systems that employ a single illumination source (projector) and a single camera, those that project sweeping laser light plane, black-and-white (BW) stripe patterns, gray-level stripes have been well investigated [1][2][3]. Since they are projecting multiple light patterns or sweeping laser stripe, they are appropriate for stationary scenes. Hall-Holt and Rusinkiewicz have recently suggested a method based on time-varying binary BW patterns for near real-time ranging in four video frames, but object motion is assumed to be slow to keep "time coherence" [4].

Various approaches have been made to one-shot or near one-shot imaging. One class of methods is those that project continuous light patterns: Tajima and Iwakawa used rainbow pattern with continuous change of light color [5], Huang *et al.* used sinusoidal color fringe light with continuous variation of phase [6], and Carrihill and Hummel used gray-level ramp and constant illumination [7]. Although all these methods can, in principle, produce range images with high speed and high resolution only restricted by system resolution, they are highly susceptible to system noise, nonlinearity and object surface colors. Another class of approaches includes those that use discrete color patterns: Davis and Nixon designed a color-dot illumination pattern, Boyer and Kak developed color stripe patterns that can be identified by a color coding of adjacent stripes, and Zhang *et al.* also developed a color stripe pattern based on a de Bruijn sequence and stripes are identified by dynamic programming [8][9][10].

Although the color-dot and color-stripe methods are less sensitive to system noise and nonlinearity compared to those with continuous light patterns, they are also significantly affected by object color reflectance and their range resolution is limited by stripe width. Caspi *et al.* presented a three-image-frame method that can overcome the ambiguity in stripe labeling due to object surface color, but its real-time application has not been explicitly considered [11].

Most of the color stripe-based methods suggest design of color patterns that can be uniquely identified in the illumination space, but little explicit attention has been paid to the selection of colors. In this paper, we investigate the selection of colors for illumination stripe patterns for maximizing range resolution, and present a novel two-shot imaging method which is insensitive to system noise, nonlinearity and object color reflectances.

The rest of this paper is organized as follows. Section 2 describes a design of color multiple-stripe pattern, Section 3 discusses selection of colors for one-shot imaging, and Section 4 presents a method for two-shot imaging. In Section 5, generation and identification of multiple stripe patterns are discussed. Section 6 presents the experimental results and Section 7 concludes this paper.

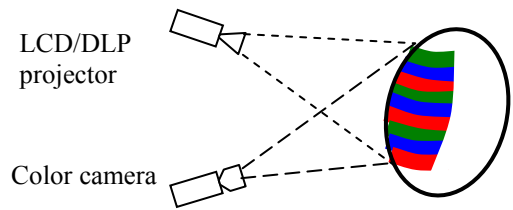

**Fig. 1.** Triangulation-based structured-light range imaging.

## 2 Multiple-Stripe Patterns for Structured Light

A typical triangulation-based ranging system with structured light consists of an LCD or DLP projector as illustrated in Fig. 1. Design of a good light pattern is critical for establishing reliable correspondences between the projected light and camera. For real-time (30 Hz) one-shot imaging, only one illumination pattern is used and fixed in time. For two-shot imaging, on the other hand, the two light patterns should alternate in time and the projector and camera should be synchronized. Since projectors and cameras with frame rates higher than 60 Hz are now common commercially, near real-time imaging with two shots becomes much more feasible than before. In this section, we describe methods for selecting stripe patterns and colors for one-shot and two-shot imaging.

The most straightforward way of generating unique color labels for $M$ stripes would be to assign $M$ different colors. In this case, the color distances between the stripes are small and this simple scheme can be as sensitive to system noise and nonlinearity as the rainbow pattern [5]. A small number of colors with substantial color differences are more desirable in this regard, but global uniqueness is hard to achieve due to the repeated appearance of a color among $M$ stripes. This problem has been addressed and *spatially windowed uniqueness* has been investigated [9][10]. Instead of using one stripe for identification, we can use a small number (e.g., $k$) of adjacent stripes such that sub-sequence of $k$ consecutive stripes is unique within the entire stripe sequence.

It can be easily shown that from $N$ different colors, $N^k$ different stripe sequences with the length $k$ can be made. When adjacent stripes are forced to have different colors, the number of uniquely identified sequences is [9]:

$$n(N,k) = N(N-1)^{k-1} \qquad (1)$$

If a single stripe is used with $N = 3$ colors, for instance, only 3 stripe labels are attainable, but the number of uniquely identifiable labels increases to 6, 12, 24 and 48 for $k = 2, 3, 4$ and 5. The entire pattern should be generated such that any sub-pattern around a stripe can be uniquely labeled.

It may be noted that with a binary BW pattern ($N = 2$) it is impossible to increase the distinct labels in this way of generating subsequences since $n(2,k) = 2 \cdot (2-1)^{k-1} = 2$. In other words, for one-shot imaging, the number of identifiable sub-patterns remains fixed regardless of the length $k$. Hall-Holt and Rusinkiewicz used multiple frames in time for increasing it with binary BW stripes [4].

If the color subpatterns are simply concatenated, not every stripe is uniquely identifiable [9]. We prefer having the entire stripe pattern designed such that the windows of subpatterns overlap but every stripe can be identified by a unique subpattern of $k$ consecutive stripes centered at the stripe. Zhang *et al.* have employed this scheme using the de Bruijn sequence [10].

The design of the stripe pattern requires several choices for: the total number of stripes $M$, the length of the subpattern $k$ and the number of colors $N$. In what follows, we turn into a discussion of criteria for choosing the stripe colors for high-resolution range imaging.

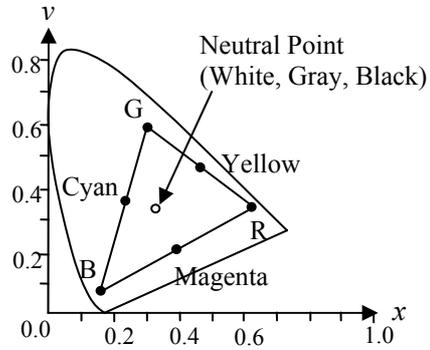

**Fig. 2.** CIE-*xy* chromaticity diagram of colors.

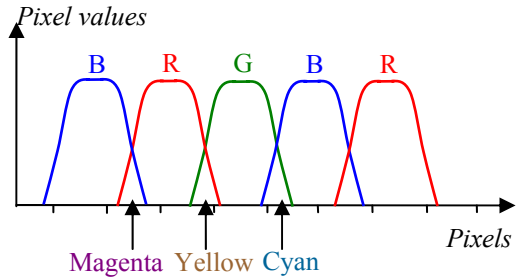

**Fig. 3.** Intensity profiles of RGB stripes.

## 3   Color Selection for One-Shot Imaging

For high-resolution ranging, the stripes should be as narrow as possible. For a given resolution limit from a set of projector and camera, stripes can be better detected with higher color contrast between the stripes. The most common choice of colors has been among R, G, B, cyan (C), yellow (Y) and magenta (M), black, and white. For the stripes that appear as thin as 1.5~2 pixels in the camera, the use of some colors confound color stripe detection.

   Let us first consider the case of using only deeply saturated three primary colors R, G and B, and addition of other colors later. Fig. 2 shows a chromaticity space (CIE-*xy* space). The colors that can be represented by the system RGB filter primaries are limited to the triangle shown in Fig. 2. For the sake of simplicity in discussion, we assume that the filter characteristics of projector and camera are identical. The image irradiance in the camera $I(\lambda)$ can be represented as:

$$I(\lambda) = g_\theta S(\lambda) E(\lambda) ,\qquad(2)$$

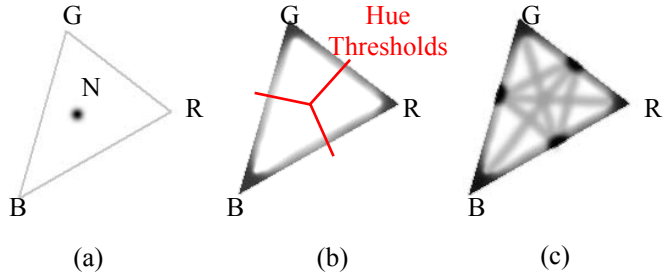

**Fig. 4.** Chromaticity diagram: (a) neutral object reflectances, (b) dispersed chromaticities under RGB stripe illumination and (c) dispersed chromaticities under RGBCMY stripe illumination.

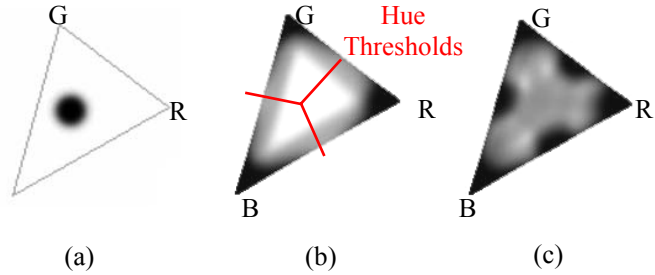

**Fig. 5.** Chromaticity diagram: (a) colored object reflectances, (b) dispersed chromaticities under RGB stripe illumination and (c) dispersed chromaticities under RGBCMY stripe illumination.

where $S(\lambda)$ is the object reflectance, $E(\lambda)$ is the illumination from the projector, and $g_\theta$ is the geometric shading factor determined by the surface orientation and illumination angle. When object reflectances $S(\lambda)$s are neutral (Fig. 4(a)), the received colors are determined by the projector illumination $E(\lambda)$, i.e., RGB stripe illumination, which splits the reflections from the neutral surfaces to the RGB points, i.e., the vertices in the chromaticity triangle (Fig. 4(b)). The one-dimensional hue value can be used effectively for separating the distinct colors. The more saturated the three stripe-illumination colors are, the farther apart the image stripe chromacities are and therefore the easier to detect against each other.

For contiguous RGB stripes, other colors than RGB appear in the image from the linear combination of RGB colors due to the limited bandwidth of the camera/projector system as illustrated in Fig. 3. The linear combinations of RG, GB and BR colors are generated at the boundaries. Fig. 4(a) shows neutral reflectance in the chromaticity triangle, and 4(b) shows the spread of the chromaticities at the boundaries. With only RGB illumination colors, thresholding by hue is effective in the presence of false boundary colors as illustrated in Fig. 4(b). Hue thresholding also works well for the reflections from moderately colored object surfaces as illustrated in Fig. 5. Fig. 5(a) and (b) depict the chromaticities of colored objects and spread chromaticities under RGB stripe illumination, respectively.

When the stripe width is 1.5~2 pixels, the pixels with the false colors around the stripe boundaries is substantial compared to the stripe colors. When additional colors such as CMY are used, the false boundary colors significantly confound stripe detection and identification since the CMY colors are linear combinations of the RGB primaries. The false colors can easily break subpatterns. Fig. 4(c) and 5(c) depict the chromaticity spreads under RGBCMY illumination from neutral and colored objects, respectively. It can be seen that additional false colors appear and there is no easy way to separate them from the stripe colors.

We find that for high-resolution imaging, the use of only RGB colors results in the best resolution and the smallest errors. This restricts the number of color code $N$ to 3. White can be added if surface shading is insignificant and black might be added when ambient light is low.

Other 3-color primaries such as CMY would perform similarly to RGB only if the chromaticity distances of the primaries were as large as those of RGB. However, the color distances between CMY (as synthesized from the linear combinations of RGB) are substantially smaller. (See Fig. 2.) The CMY colors in Fig. 2 have substantially less saturation than RGB. The only way to keep the CMY distances comparable to those of RGB is to employ narrowband CMY color filters in the camera instead of RGB, but such commercial cameras are rare in reality.

If object surfaces have substantially saturated colors, the reflectance chromaticities are dispersed so widely that even strong RGB stripes cannot separate them for detection and extra information is needed.

## 4  Two-Shot Imaging Method

When synchronized high-speed camera and projector are available, more than one frame can be used to discount the effects of highly saturated object colors and ambient illumination in range sensing. Many commercial cameras and projectors offer external trigger and frame refresh rate higher than 60 Hz. We present a two-shot imaging method that uses two video frames and stripes with highly saturated projector colors.

When projection of two light colors $E_1(\lambda)$ and $E_2(\lambda)$ alternates in time, the following two images can be obtained:

$$\begin{cases} I_1(\lambda) = g_\theta S(\lambda)[E_1(\lambda) + A(\lambda)] \\ I_2(\lambda) = g_\theta S(\lambda)[E_2(\lambda) + A(\lambda)] \end{cases} \quad (3)$$

where the effect of ambient illumination is included. Since it is assumed that objects are stationary during two consecutive video frames in a short time interval, $g_\theta$ and $S(\lambda)$ are common to both the images.

Caspi *et al.* used an extra image with the projector turned off to estimate the influence of the ambient illumination from $I_A(\lambda) = g_\theta S(\lambda) A(\lambda)$ [11]. After discounting $A(\lambda)$, the ratio of the images will be dependent only on the designed illumination colors without any influence of surface color and shading, i.e.,:

$$\frac{I_2(\lambda)-I_A(\lambda)}{I_1(\lambda)-I_A(\lambda)} = \frac{E_2(\lambda)}{E_1(\lambda)}. \tag{4}$$

With the commonly used assumption of spectral smoothness of $S(\lambda)$ in each color channel and some appropriate color calibration described in [11], the responses in the color channels can be decoupled and analyzed independently with the following ratios:

$$\frac{R_2}{R_1}, \quad \frac{G_2}{G_1} \quad \text{and} \quad \frac{B_2}{B_1}.$$

While this is an effective way of discounting objects colors if image values are in the linear range of projector and camera responses and many combinations of color ratios can be produced for stripes, the ratios become unstable when $I_1$ and $I_2$ are small due to small $g_\theta$ and $S(\lambda)$ values and when image values are clipped on highly reflective surfaces. The geometric factor $g_\theta$ is small on shaded surfaces whose surface orientation is away from the illumination angle. A "three-shot" method by Caspi *et al.* uses an extra shot to measure the ambient illumination $I_A(\lambda)$ for its compensation and relies on the color ratios [11]. However, the color ratios are highly unstable for bright and dark regions.

Instead of assigning many RGB projector colors and identifying them in the linear range, we seek a small number of stripe signatures from the sign of color difference to reduce the sensitivity to light intensity and ambient light without the third image for the estimation of ambient light. From Equation 3, the difference of two images is given as:

$$\begin{aligned} \Delta I(\lambda) &= I_2(\lambda) - I_1(\lambda) \\ &= g_\theta S(\lambda)[E_2(\lambda) - E_1(\lambda)], \\ &= g_\theta S(\lambda)\Delta E(\lambda) \end{aligned} \tag{5}$$

and the ambient illumination is discounted. In Equation 5, it can be seen that though $\Delta I(\lambda)$ is significantly affected by $g_\theta$ and $S(\lambda)$, its sign is not since $g_\theta$ and $S(\lambda)$ are both positive. We can derive a few stripe labels from the sign of image difference.

When color channels are decoupled with the same color assumption described above, the differences of RGB channel values are:

$$\begin{cases} \Delta R = g_\theta S^R \Delta E^R \\ \Delta G = g_\theta S^G \Delta E^G \\ \Delta B = g_\theta S^B \Delta E^B \end{cases}, \tag{6}$$

where $S^R$, $S^G$ and $S^B$ are the object reflectances in R, G and B channels, respectively, and $\Delta E^R$, $\Delta E^G$ and $\Delta E^B$ are the intensity differences of projection illumination in R, G and B channels, respectively. From the positive and negative signs of $\Delta R$, $\Delta G$ and $\Delta B$, we can obtain $2^3=8$ distinct codes to assign in a projector stripe. If we construct subpatterns with $N=8$, $n(N,k) = 8\cdot 7^{(k-1)}$ unique subpatterns can be generated according to

Equation 1. In the design of subpatterns, however, it was observed that the spatial transition of colors over the stripes in multiple channels make false colors due to the inconsistency of color channels and those false colors are not easily identifiable. If we allow spatial color transition only in one of the RGB channels, only three types of transitions are allowed from one stripe to another. In this case, it can be shown that the number of unique subpatterns for the length $k$ in two-shot, "$n2$" is given as:

$$n2(m,k) = 2^m m^{k-1} \qquad (7)$$

where $m$ is the number of possible spatial color transitions, e.g., $m=3$ for RGB colors and $m=2$ for GB colors. To maximize the $\Delta R$, $\Delta G$ and $\Delta B$ values for good discriminability between the positive and negative signs, the intensity difference of projection color should be maximized; we assign the minimum and maximum programmable values for the frame 1 and 2.

The channel value differences in Equation 6 are also affected by small values of $g_\theta$ and $S(\lambda)$, but we claim that their influence is much smaller with the difference than the ratio for given system noise. Furthermore, the effect of system nonlinearity and RGB color clipping is much less pronounced with the signs of the difference. To demonstrate the efficacy of this approach, we use only presented method in our experiments with only two color channels, G and B, in our experiments, i.e., with $N=2^2=4$ codes.

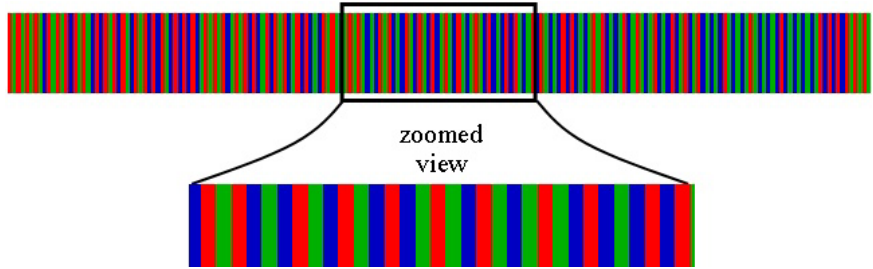

**Fig. 6.** RGB pattern for one-shot imaging: 192 unique subpatterns.

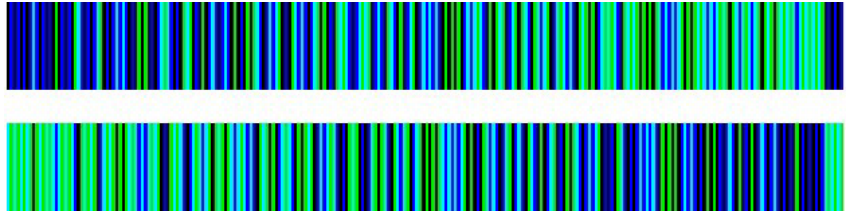

**Fig. 7.** GB patterns for two-shot imaging: 256 unique subpatterns in each frame.

## 5 Multiple-Stripe Pattern Synthesis and Identification

The requirements for a program that generates the subpatterns are as follows:

(a) Different colors or codes should be assigned to adjacent stripes to make distinguishable stripes.
(b) The subpattern generated by the $i$th stripe should be different from any subpatterns generated up to the ($i$-1)th stripe.

For the two-shot imaging, the following extra requirements should be added:

(c) Only one channel among RGB should make a color transition between the adjacent stripes.
(d) Colors in each stripe in the second frame should be the reverse of those in the first frame in each channel. The reverse of maximum value is the minimum value and *vice versa*.

There is a tradeoff to make between the number of total stripes to cover a scene $M$ and the number of stripes in a whole pattern $n(N, k)$. For high-resolution imaging $M$ should be kept large and $n(N, k)$ should be close to $M$ for a unique encoding of all the stripes. Otherwise, the whole pattern should be repeated for the scene, and its subpatterns are not globally unique. This means that for small $N$, the length of the subpattern $k$ should be large. Wide subpatterns, however, are not reliable near the object boundaries and occlusions. The best compromise we make for one shot imaging ("$n1$") is to have $n1(3, 7)=192$ with $k=7$ for $M\cong 400$ and let the pattern appear twice. Stripe identification or unwrapping is not difficult since identical subpatterns appear only twice and they are far apart in the image. For the two-shot imaging with only GB channels, $n2(2, 7)= 2^2 \cdot 2^{7-1} =256$ with $k=7$. The generated patterns for one-shot and two-shot image are shown in Figs. 6 and 7.

Stripe segmentation in the received image can be done mostly based on two methods: stripe color classification and edge detection by color gradients. Both should work equivalently for imaging with thick stripes. For high frequency stripe imaging, however, color gradients are not stably detected after smoothing and we prefer stripe color classification.

## 6 Experimental Results

We have carried out experiments with a number of objects using the described one-shot and two-shot methods. We used a Sony XC-003 3-CCD 640×480 color camera, and an Epson EMP-7700 1024×768 projector. The encoded patterns are projected onto the various objects by the projector, and the camera captures the scene. We did not use any optimization algorithm for extracting stripe identification labels since the direct color classification and decoding algorithms work well. The projected patterns all consist of stripes with one-pixel width, and the width of stripes in the captured images is around 2 pixels. We used RGB stripes with $k=7$ and $n1=192$ for one-shot imaging and GB stripes with $k=7$ and $n2=256$ for two-shot imaging, as described in Section 5. We used only 2 color channels because they keep $n2$ large enough, and the reason why we chose GB instead of RG is that the G and B color channels have slightly less crosstalk than the R and G channels in our camera-projector setup.

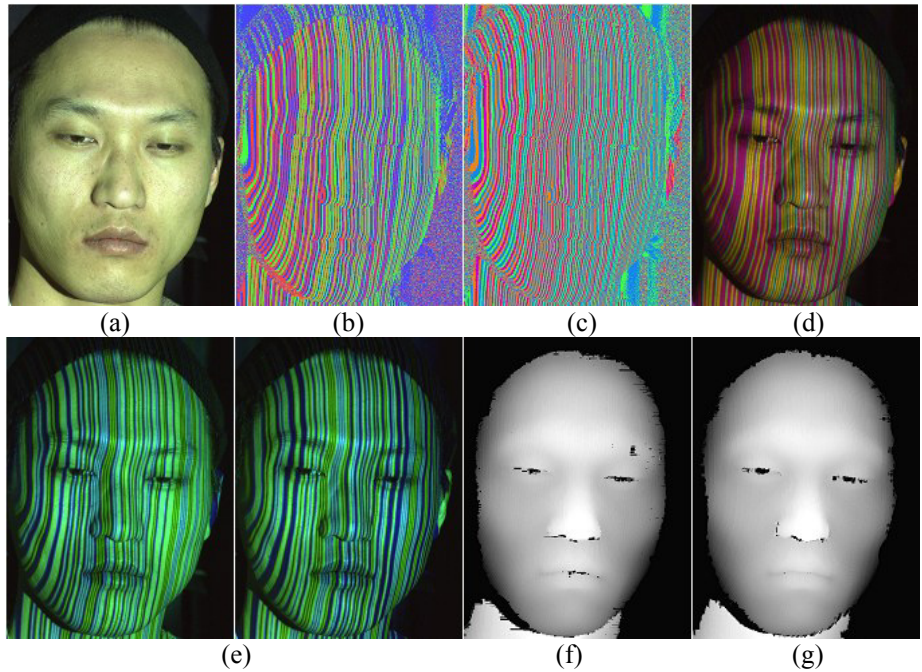

**Fig. 8.** Results from a human face: (a) A human face under white light, (b) identified stripes from subpattern from one-shot imaging (pseudo color assignment for each stripe), (c) from two-shot imaging (pseudo color assignment for each stripe), (d) color stripes for one-shot imaging, (e) two stripe patterns for two shot imaging, (f) range image from one-shot imaging, and (g) range image from two-shot imaging.

Fig. 8 shows the results from a human face with one-shot and two-shot imaging. For one-shot imaging, Figs. 8(a), (d), (b) and (f) show the subject under white projector illumination, stripe pattern projection, pseudo-color display of identified stripes from subpatterns, and the range image from the identified stripes, respectively. For two-shot imaging, Fig. 8(e), (c) and (g) show the two stripe patterns, pseudo-color display of identified stripes from subpatterns, and the range image from the identified stripes, respectively. Since the face has moderate colors, both the methods work well.

One-shot and two-shot imaging has also been tested with a flat color panel with highly saturated color patches. Fig. 9(a), (b), (c) and (d) shows the color panel under white light, one of the color patterns for two-shot imaging, the range image from one-shot imaging, and the range image from two-shot imaging, respectively. It can be seen that the strong colors of surface reflectance confound the one-shot imaging sig-nificantly. Fig. 10(a), (b), (c) and (d) show chromaticity plot from the human face, those from the color panel under white light, and from RGB stripe light, and the plot of panel colors in GB space from two-shot imaging.

As can be seen from Fig. 10(a), the face colors are moderate and good for the one-shot imaging. However, the colors from the panel are so strong that projection of RGB colors cannot reliably separate the colors of stripe-like regions on the captured image into the different regions in the chromaticity as shown in (b) and (c), while we can easily identify GB-combination codes in (d).

Fig. 11 shows (a) a strongly colored object (Rubik's cube), (b) its stripe segmentation and (c) the high-resolution range result using the presented two-shot method. The stripes are properly segmented in spite of the strong surface colors and specular reflections, and even so is the black paper region at the bottom. Note that the range result is sufficiently good despite the discontinuities of the multiple-stripe sequence in the black lattice regions. Fig. 12 compares the result from a colored cylinder with the proposed two-shot method and that with the method of Caspi *et al.* in [11]. It can be seen that the presented method has the advantage in the bright (or highly saturated) and dark regions.

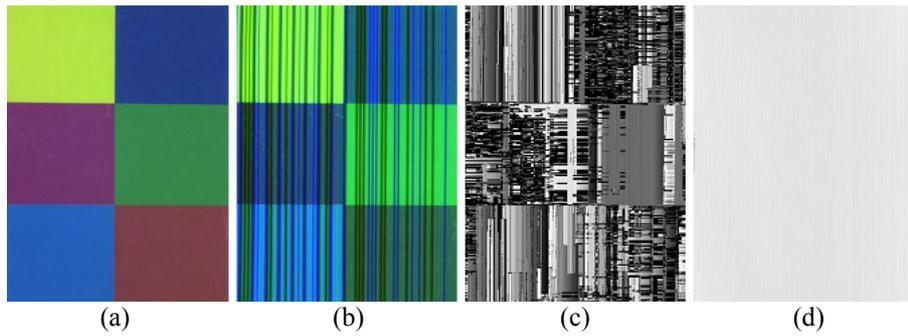

(a) (b) (c) (d)

**Fig. 9.** Experiments with a color panel: (a) under white light, (b) one of the color patterns for two-shot imaging, (c) range image from one-shot imaging, and (d) range image from two-shot imaging.

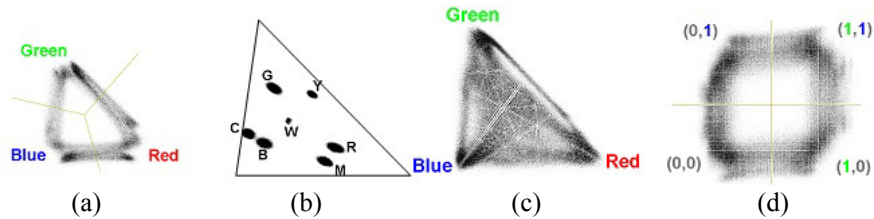

(a) (b) (c) (d)

**Fig. 10.** Color plots: (a) Chromaticities from the human skin under RGB stripe light, (b) those from the color panel under white light, and (c) those from RGB stripe light, and (d) plot of panel colors in GB space from two-shot imaging.

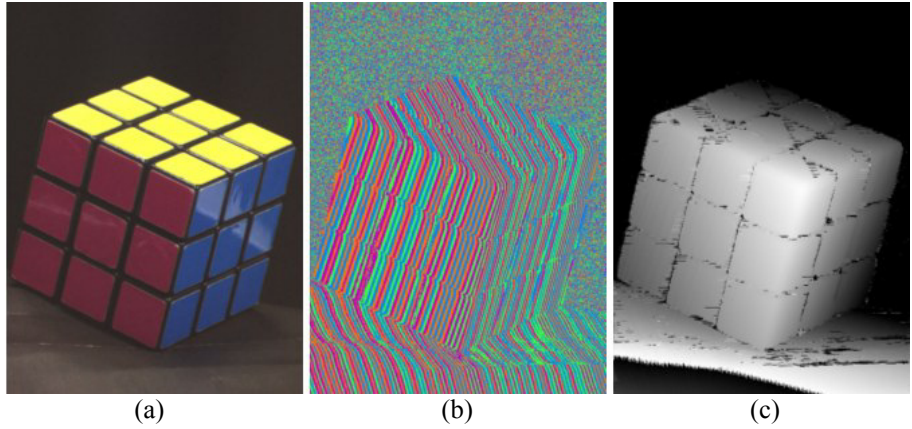

**Fig. 11.** Two-shot imaging with Rubik's cube (yellow, nearly-red and blue surfaces): (a) the object, (b) stripe segmentation by GB differences and (c) its range result.

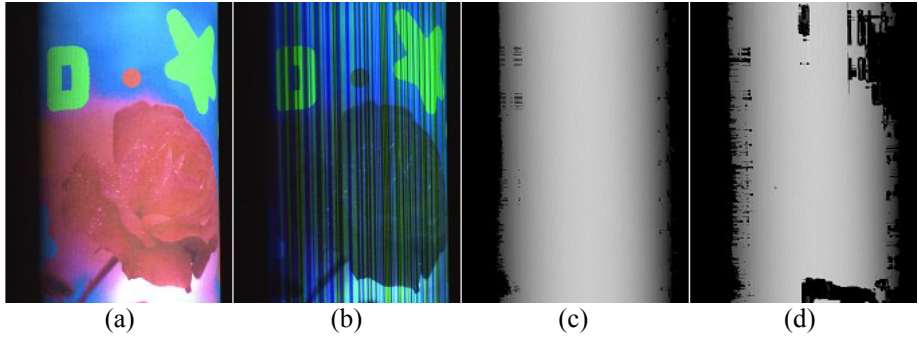

**Fig. 12.** Two-shot imaging with a cylindrical object: (a) under white illumination, (b) one of the two stripe patterns, (c) range image with the presented method, and (d) with the method in [11].

## 7 Conclusion

For rapid range sensing, we described a design of multiple-stripe patterns for increasing the number of distinguishable stripe codes, discussed a color selection scheme for reducing the ambiguity in stripe labeling in one-shot imaging, and presented a novel method for two-shot imaging that is insensitive to object color reflectance, ambient light and limitations in projector/sensor resolution. We showed that maximizing color contrast between the stripes in one-shot imaging reduce the ambiguities resulting from system resolution and object colors to some degree, and the new method of utilizing color differences in two shot imaging further reduce the ambiguities resulting from colored object surfaces, ambient light and sensor/projector noise and nonlinearity. By using the signs of color differences instead of color ratios in two-shot im-aging, we can obtain more reliable information in the bright, dark, and strongly-colored regions of objects and also minimize the number of shots in multiple-frame imaging.

## Acknowledgement

This work was supported by grant number R01-2002-000-00472-0 from the Basic Research program of Korea Science & Engineering Foundation (KOSEF).